\documentclass[10pt,twocolumn,letterpaper]{article}

\usepackage{wacv_25}
\usepackage{times}
\usepackage{epsfig}
\usepackage{graphicx}
\usepackage{subcaption}
\usepackage{comment}
\usepackage{amsmath}
\usepackage{amssymb}
\usepackage{booktabs}
\usepackage{multirow}
\usepackage{tabularx}
\usepackage{float}
\floatstyle{plaintop}
\restylefloat{table}
\usepackage{placeins}
\restylefloat{table}
\usepackage{booktabs}

\usepackage[tableposition=top]{caption}

\usepackage[capitalize]{cleveref}
\crefname{section}{Sec.}{Secs.}
\Crefname{section}{Section}{Sections}
\Crefname{table}{Table}{Tables}
\crefname{table}{Tab.}{Tabs.}

\usepackage{array} 
\newcolumntype{Z}{>{\centering\arraybackslash}p{1.3cm}}

\def\wacvPaperID{*****} 
\def\confName{WACV}
\def\confYear{2025}

\begin{document}

\title{InsertDiffusion: Identity Preserving Visualization of Objects through a Training-Free Diffusion Architecture}

\author{Phillip Mueller\\
BMW Group\\
Knorrstrasse 147, 80788 Munich, Germany\\
{\tt\small phillip.mueller@bmw.de}
\and
Jannik Wiese\\
BMW Group\\
Knorrstrasse 147, 80788 Munich, Germany\\
{\tt\small jannik.wiese@bmw.de}
\and
Ioan-Daniel Craciun\\
Technical University of Munich\\
Boltzmannstraße 15, 85748 Garching, Germany\\
{\tt\small ioandaniel.craciun@gmail.com}
\and
Lars Mikelsons\\
University of Augsburg\\
Am Technologiezentrum 8, 86159 Augsburg, Germany\\
{\tt\small lars.mikelsons@uni-a.de}
}

\maketitle
\thispagestyle{empty}

\begin{figure*}[t]
\centering
\includegraphics[width=\linewidth]{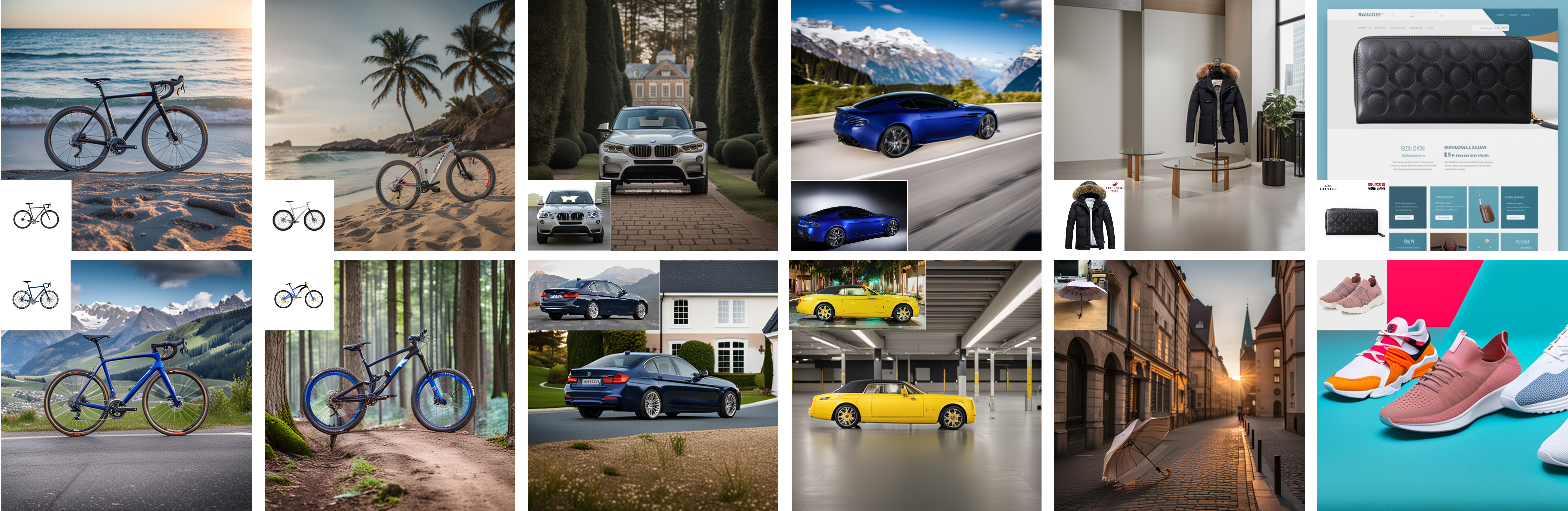}
\caption{{\bf Realistic object representations in existing and generated backgrounds} without the necessity for training or finetuning any parts of the architecture.}
\label{fig:Insert_Examples}
\vspace*{-0.25cm}
\end{figure*}

\begin{abstract}
Recent advancements in image synthesis are fueled by the advent of large-scale diffusion models. Yet, integrating realistic object visualizations seamlessly into new or existing backgrounds without extensive training remains a challenge. This paper introduces InsertDiffusion, a novel, training-free diffusion architecture that efficiently embeds objects into images while preserving their structural and identity characteristics. Our approach utilizes off-the-shelf generative models and eliminates the need for fine-tuning, making it ideal for rapid and adaptable visualizations in product design and marketing. We demonstrate superior performance over existing methods in terms of image realism and alignment with input conditions. By decomposing the generation task into independent steps, InsertDiffusion offers a scalable solution that extends the capabilities of diffusion models for practical applications, achieving high-quality visualizations that maintain the authenticity of the original objects.
\end{abstract}

\section{Introduction}

Image generation is witnessing remarkable advancements with the rise of diffusion models, achieving unprecedented levels of realism and naturalness in synthetic images ~\cite{hoDenoisingDiffusionProbabilistic2020, songDenoisingDiffusionImplicit2022, dhariwalDiffusionModelsBeat2021, sahariaPhotorealisticTexttoImageDiffusion2022}. The evolution of latent diffusion models, especially Stable Diffusion (SD) \cite{rombachHighResolutionImageSynthesis2022} and its variants such as Stable Diffusion XL (SDXL) \cite{podellSDXLImprovingLatent2023}, further improve generalization, quality, and realism and allow for a variety of conditioning mechanisms such as text and reference images. A crucial component in these advancements is the development of CLIP \cite{radfordLearningTransferableVisual2021}, which provides a  foundation for referencing text to visual concepts. Due to its generative capabilities and adaptability, SD sparked a wave of subsequent modifications and extensions to further increase the levels of image quality, customization and user-control.

Examples to condition the generation include sketches and spatial maps \cite{zhangAddingConditionalControl2023}, shape-based guidance \cite{parkShapeGuidedDiffusionInsideOutside2023}, as well as semantic segmentations and keypoints \cite{mouT2IAdapterLearningAdapters2023}. Image editing has also seen significant progress. Besides text-driven image editing \cite{wangPretrainingAllYou2022, andonianPaintWord2023, tumanyanPlugandPlayDiffusionFeatures2022}, point-based dragging approaches \cite{shiDragDiffusionHarnessingDiffusion2023, mouDragonDiffusionEnablingDragstyle2023} and inpainting methods \cite{lugmayrRePaintInpaintingUsing2022} extend the capabilities of diffusion models for image editing.

In this paper, we investigate the task of realistic object visualization, which involves injecting a given object into an existing or newly generated background and merging both representations to create a perceptually appealing scene while maintaining the object's structure and characteristics. Some examples are visualized in \cref{fig:Insert_Examples}. This task is particularly relevant for applications in product and engineering design, as well as customer-oriented marketing. Proposed applications include rendering geometric or CAD-like objects as realistic images and enabling customization in advertising and personalization (e.g.: visualizing a new car or bike in a customer's driveway). We specifically aim to visualize technical representations of products like bicycles as well as design-representations of consumer-products like cars. 

Our approach\footnote{Code is found under: \url{https://github.com/ragor114/InsertDiffusion}} aims to decouple the highly customized models and workflows required for generating technically accurate images from the visualization and scenic representation. By leveraging publicly available, large-scale diffusion models, we can visualize the results in realistic scenes without the need for training or fine-tuning. This allows smaller, domain-specific models to focus on generating specific types of images while utilizing the extensive capabilities of models trained on millions of images for realistic rendering. 

Previous studies have explored this task from various angles, but none have addressed it from the perspective of fully leveraging existing capabilities in foundation models as straightforward as possible while utilizing publicly-available implementations only and avoiding training or finetuning altogether. Methods such as TF-Icon \cite{luTFICONDiffusionBasedTrainingFree2023}, AnyDoor \cite{chenAnyDoorZeroshotObjectlevel2023}, and PrimeComposer \cite{wangPrimeComposerFasterProgressively2024} aim to inject objects into given backgrounds. However, TF-Icon and PrimeComposer, which are both training-free, modify the injected object to align with the background style, altering its characteristics. Both approaches rely on extracting, modifying and reinjecting attention maps in a non-straightforward fashion. AnyDoor learns detail- and ID-extractors to achieve this goal and is, therefore, not training-free. Another comparable method, CollageDiffusion \cite{sarukkaiCollageDiffusion2023} merges multiple images into one collage but is also not training-free. 

For realistic image insertion into newly generated backgrounds, the state-of-the-art methods ReplaceAnything \cite{chenReplaceAnythingYouWant2024} and ObjectDrop \cite{winterObjectDropBootstrappingCounterfactuals2024} have been proposed. ReplaceAnything is based on an architecture that generates realistic images of humans interacting with products \cite{chenVirtualModelGeneratingObjectIDretentive2024} that has to be trained. ObjectDrop finetunes a diffusion model on a dataset containing counterfactual images to modify images by inserting, removing, or moving objects.

Despite the remarkable results that the existing methods achieve, we analyze that they do not fully leverage the inherent capabilities of large-scale image generation models. Numerous studies have demonstrated the extensive capabilities of SD in other contexts \cite{poStateArtDiffusion2023, tangEmergentCorrespondenceImage2023, luoDiffusionHyperfeaturesSearching2023}. To ensure consistency, adaptability, and ease of use, we propose a significantly simpler method that utilizes off-the-shelf generative models available through the diffusers-library on HuggingFace \cite{vonplatenDiffusersStateoftheartDiffusion2024} for all tasks. Our architecture is designed to be adaptable to the fast-evolving field of diffusion models, allowing for the replacement of any component in the architecture as new, improved versions become available. 

In essence, we create a mask from the object and pass it, along with the object, to SD using the inpainting function. The inverse of the object mask defines the area in the background-image that the model can modify, while the object itself remains unchanged. After generating an intermediate image composition, we apply an image-to-image transformation that noises and then denoises the composed image again to optimize high-frequency structures.

\section{Related Work}
\subsection{Image-to-Image Transformation}
Image-to-image transformations include a variety of tasks like local image editing, colorization, inpainting, uncropping, upscaling, and style changes.
Whereby colorization, style-transfer, inpainting, uncropping, and upscaling can be seen as subtasks of object insertion.
Tumanyan et al. \cite{tumanyanPlugandPlayDiffusionFeatures2022} manipulate the spatial features and self-attention layers of a pretrained SD model during the generation process. They inject features from the initial image into the text-guided image generation. Palette \cite{sahariaPaletteImagetoImageDiffusion2022} pursues a different approach in proposing a unified framework for image-to-image translations using conditional diffusion models. The input image is noised and then iteratively denoised. The denoising process starts at an intermediate, noisy representation of the input image and is conditioned on text or other modalities. For latent diffusion models like SD, the input image is encoded and noise is added to the latent representation. The image-to-image implementation in Diffusers is based on SD and therefore also operates with latent images \cite{vonplatenDiffusersStateoftheartDiffusion2024}.

\subsection{Inpainting}
\begin{figure}[h]
\centering
\includegraphics[width=\linewidth]{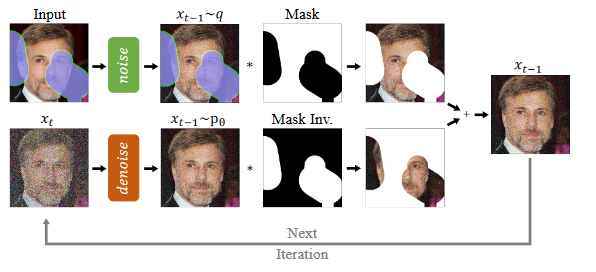}
\caption{{Local image editing through inpainting as proposed in RePaint \cite{lugmayrRePaintInpaintingUsing2022}}}
\label{fig:RePaint}
\vspace*{-0.25cm}
\end{figure}

Inpainting is a common method for local image editing. It relies on a mask to determine which regions in an image can be modified by the diffusion model. In each step in the generation, the initial image is noised according to the current timestep. Its unmasked regions are merged with the masked regions modified by the diffusion model and forwarded into the next denoising timestep \cite{lugmayrRePaintInpaintingUsing2022}. This process is shown in \cref{fig:RePaint}. Inpainting functionalities based on SD are provided in Diffusers \cite{rombachHighResolutionImageSynthesis2022, vonplatenDiffusersStateoftheartDiffusion2024}.

\subsection{Object Insertion}
Most existing works for object insertion employ finetuning or training of an additional adapter. AnyDoor \cite{chenAnyDoorZeroshotObjectlevel2023} is designed for zero-shot "object teleportation" into a given scene at specifiable locations by utilizing identity features from the target image and detail features of the target-scene composition. The identity features are extracted using a finetuned, self-supervised visual encoder (DINO-V2 \cite{oquabDINOv2LearningRobust2024}). The detailed features are represented using high-frequency maps that capture fine image details while allowing local variations. These are then mapped into the diffusion U-Net by a finetuned, ControlNet-style encoder \cite{zhangAddingConditionalControl2023}. 

ObjectDrop \cite{winterObjectDropBootstrappingCounterfactuals2024} leverages a dataset of "counterfactual" pairs of images that show the scene before and after object removal to finetune SD for object removal. This model is used to synthetically create a larger dataset of counterfactual image pairs and subsequently finetune SD for object insertion.
PrimeComposer \cite{wangPrimeComposerFasterProgressively2024} steers the attention weights at different noise levels to preserve the object appearance while composing it with the background in a natural way. Additionally, they employ Classifier-Free Guidance \cite{hoClassifierFreeDiffusionGuidance2022} to enhance the quality of the composed images.
Paint-by-Example \cite{yangPaintExampleExemplarbased2022} leverage SD \cite{rombachHighResolutionImageSynthesis2022} and Classifier-Free Guidance \cite{hoClassifierFreeDiffusionGuidance2022} and employ self-supervised training to disentangle and re-organize the background image and the object. 

TF-Icon \cite{luTFICONDiffusionBasedTrainingFree2023} is a training-free method for cross-domain image compositions. They invert the real images into latent codes using an exceptional prompt that contains no information. The latent codes are then used as a starting point for the text-guided image generation process. Composite self-attention maps are injected to infuse contextual information from the background image into the inserted object.

Shopify-Background-Replacement (SBR) \cite{shopifyShopifyImageBackground2023}, first extracts the object from the original background by using a depth estimation model, taking only the foreground. Then the depth image and the text prompt are passed to SDXL-Turbo \cite{sauerAdversarialDiffusionDistillation2023} augmented by a ControlNet \cite{zhangAddingConditionalControl2023} which handles the depth map. After inferring a new background the original extracted foreground is pasted on top of the background to generate the final image.

For the task of background replacement, Chen et al. \cite{chenReplaceAnythingYouWant2024} present ReplaceAnything. This state-of-the-art method is based on their previous work  VirtualModel \cite{chenVirtualModelGeneratingObjectIDretentive2024} which visualizes consumer products in new backgrounds as if they were held by a human model. The trained VirtualModel consists of a Content-guided branch to ensure the consistency of the product, and a Interaction-guided branch to guide the model in creating realistic product-human interactions. ReplaceAnything is currently only available as a HuggingFace demo.

\section{Method}

\begin{figure*}[h]
\centering
\includegraphics[width=\linewidth]{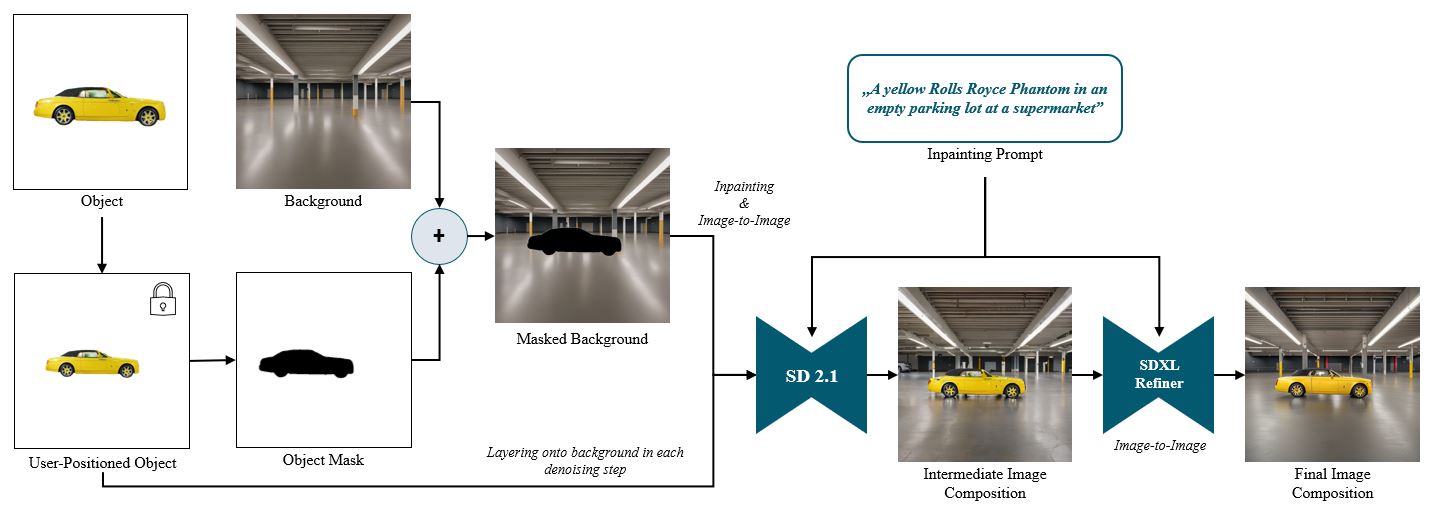}
\caption{The {\bf InsertDiffusion Architecture} is designed to seamlessly insert an object into a new background while preserving the geometry and key visual characteristics of the object. After the object is scaled and positioned by the user, an object-mask is created automatically and composed with the background image. The masked background is passed to SD together with the original object image. Using the image-to-image and inpainting functions, the original image is layered onto the background for each denoising step. The resulting intermediate image composition is subsequently refined by a second diffusion model (SDXL).}
\label{fig:Architecture}
\vspace*{-0.25cm}
\end{figure*}

The architecture of InsertDiffusion is shown in \cref{fig:Architecture}. The main idea is to modify image characteristics, such as shadows, lighting and texture in both object and background to obtain a realistic composition. This is done without modifying the attention layers, training an additional adapter or finetuning any part of the diffusion model. Given an object to insert, the user can either use an existing background image or generate a new background with SDXL \cite{podellSDXLImprovingLatent2023}. Once the object is isolated from its original background, we create a mask to "reserve" the desired location in the new background image and pass both the mask and the original object to the background refinement. This is done to adapt the background such that it seamlessly incorporates the object. In a second step, we refine the intermediate composition of object and background by adding some noise and subsequently denoising again.

\subsection{Core Architecture}
Given an isolated and user-positioned object on a white background $x^{(obj)}$, the object mask $m = mask(x^{(obj)})$ is obtained from $x^{(obj)}$ by applying a threshold i.e. by setting all pixels brighter than the threshold to 0 and all others to 1.

{\bf Intermediate Image Composition.} The intermediate image composition produces a modified version of the background. In general, it can be computed by:
\begin{equation}
    \label{Intermed_Comp}
    \begin{aligned}
        \hat{z}^{(comp)}&=m\odot{z^{(obj)}}\\
        &+(1-m)\odot{G(z^{(bg)},m, z^{(obj)}, \tau_{\theta}(y)).}
    \end{aligned}
\end{equation}
Whereby, $z^{(obj)}$ is a latent representation of the object image obtained by passing it through the SD encoder, $z^{(bg)}$ is a latent representation of the background image obtained using the SD encoder and $G$ is a masked diffusion process.
The term $m\odot{z}$ ensures that the object area is preserved in the latent representation.
$(1-m)\odot{G(z^{(bg)},m, z^{(obj)}, \tau_{\theta}(y))}$ updates the background image in the regions where $m=0$. The generation is guided by the masked object and the CLIP-encoded text-prompt $\tau_{\theta}(y)$.
To iteratively refine the background and allow for seamless object integration, the latent representation of the original object $z^{(obj)}$ is injected into the background for every denoising step. This is done to keep the object itself mostly unmodified. The update for each timestep, using the latent diffusion model $\epsilon_{\theta}$, is:
\begin{equation}
    \label{Update}
    \begin{aligned}
        &\hat{z}^{(comp)}_{t-1} = m\odot z^{(obj)}_{t-1}+\\
         &+(1-m)\odot (\hat{z}^{(comp)}_{t}-\epsilon_{\theta}(\hat{z}^{(comp)}_{t},\tau_{\theta}(y),t)).
    \end{aligned}
\end{equation}
Hereby, $z^{(obj)}_{t}$ is calculated by adding noise to the latent representation of the object according to the noise schedule and timestep.
Noise is added according to the canonical formulation of the diffusion forward process \cite{hoDenoisingDiffusionProbabilistic2020} given by
\begin{equation} \label{eq:forward_diffusion}
    \begin{aligned}
    x_t &= \sqrt{\Bar{\alpha_t}}x_0 + \epsilon \sqrt{1- \Bar{\alpha}_t},\\
    &\epsilon \sim \mathcal{N}(0, \mathbf{I}), \ \alpha_t = 1 - \beta_t,\ \Bar{\alpha}_t = \prod^t_{s=0} \alpha_s
    \end{aligned}
\end{equation}
where all $\beta_t$ are defined by the noise schedule. We use the default noise schedule for each model as set in the diffusers library \cite{vonplatenDiffusersStateoftheartDiffusion2024}.
The masked diffusion process $G$ is obtained by iteratively applying Equation \cref{Update} from some initial timestep to $t=0$.
To insert the object into a given background we set the initialize $t=n$ and obtain $\hat{z}^{(comp)}_{n}$ by applying Equation \cref{eq:forward_diffusion} on a pasted composition $z^{(pasted)} = m \odot z^{(obj)} + (1-m) \odot z^{(bg)}$.
To generate a completely new background we set $t=n=T$ which is equivalent to initializing $\hat{z}^{(comp)}_{T}$ from gaussian noise and iteratively applying Equation \cref{Update}.

{\bf Refinement.} The second step in the architecture aims to refine the composed intermediate image by making it more conistent and modifying high-frequency image characteristics. The intermediate image composition $\hat{z}^{(comp)}$ obtained using Equation \cref{Intermed_Comp} is noised for $n$ timesteps using Equation \cref{eq:forward_diffusion} to obtain $z_n$. Subsequent denoising steps are again guided by the text-prompt and iteratively computed as follows:
\begin{equation}
    \label{ReDiffusion}
    z_{t-1}=z_t-\epsilon_{\theta}(z_t,t,\tau_{\theta}(y)) \text{, with}\ t=t_n,t_{n-1},…,t_0.
\end{equation}

\subsection{Optional Additions}
To make our architecture more accessible and flexible to use, we provide additional functionalities to prepare the object for background insertion. We again only utilize existing models that are publicly available within Diffusers on HuggingFace \cite{vonplatenDiffusersStateoftheartDiffusion2024} and a publically available implementation of Language Segment Anything (langSAM)\cite{kirillovSegmentAnything2023, medeirosLangSegmentAnything2023}.

{\bf Background Generation.} If no existing background is provided to insert the object, the user can also utilize existing text-to-image models like SD \cite{rombachHighResolutionImageSynthesis2022} or SDXL \cite{podellSDXLImprovingLatent2023}. After generating and choosing a suitable background the user can position the object in the background as needed and use our insertion into a given background pipeline. The background generation is simply conditioned on a background-prompt.

{\bf Colorization.} Technical images often come as drawing- or CAD-like black-and-white representations. In our experiments, we have observed that injecting such images severely deprecates the quality of the resulting image composition. Therefore, we provide a simple colorization scheme, again based on masked inpainting with SD. For low-resolution images, we observe that the coloring scheme produces low-quality results. We therefore increase the resolution by employing the upscaling version of SD. For an input image of the resolution $256 \times 256$ we upscale it by a factor of $4$ to $1024 \times 1024$. The upscaled image, a mask and a text prompt are then passed to SDXL to perform colorization. 

\begin{figure}[h]
\centering
\includegraphics[width=\linewidth]{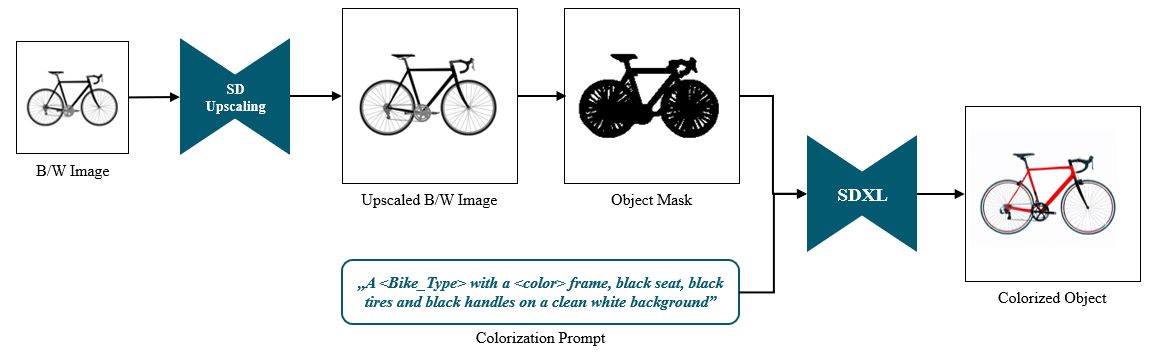}
\caption{{\bf Image Colorization scheme} for black-and-white images. Given a mask of the object, SDXL \cite{podellSDXLImprovingLatent2023} is prompted to color the object defined by the masked area. If the original image containing the object is of low resolution, we advise upscaling the object by using functionality provided by Stable Diffusion.}
\label{fig:Colorization}
\vspace*{-0.25cm}
\end{figure}

{\bf Prompting} After conducting several experiments with different prompting schemes we derive a text prompt template for our method. Using the template we only require the product-type, color, and place to be filled in by the user. We provide templates for colorization and object insertion.

{\bf Object Segmentation.} Since not all object images are already separated from their original background, we also include automatic object segmentation. For this task, we utilize langSAM \cite{kirillovSegmentAnything2023}. It only requires the specification of the object category by the user. We use the GitHub implementation by the authors, but the model is also available in the transformers-library on HuggingFace \cite{wolf-etal-2020-transformers}.

\subsection{Implementation Details}
Our model architecture is set up such that the components can be updated and replaced when more powerful models are released. For Intermediate Image Composition, we use SD-2.1 \cite{rombachHighResolutionImageSynthesis2022} as it provides stable image-to-image and inpainting capabilities. To generate the intermediate composition of the object and the background, we use 75 diffusion steps with a prompt-guidance strength of 15. 
For the second stage in the architecture, we use SDXL \cite{podellSDXLImprovingLatent2023} with its image-to-image implementation. The prompt-guidance strength is 7.5 and we noise the intermediate image for 10 out of 50 steps before denoising it again. With the scaled linear scheduler, this corresponds to the image being noised by $\sim 20\%$.
For colorization, we use SDXL \cite{podellSDXLImprovingLatent2023} for a total of 30 steps with an image-to-image strength of $91\%$, and a prompt-guidance strength of 17. The upscaling is done using SD-1.5 \cite{rombachHighResolutionImageSynthesis2022}.

\section{Experiments} \label{experiments}
Our experiments are twofold. For object injection into an existing background, we compare TF-Icon \cite{luTFICONDiffusionBasedTrainingFree2023}, AnyDoor \cite{chenAnyDoorZeroshotObjectlevel2023} and our method. ObjectDrop \cite{winterObjectDropBootstrappingCounterfactuals2024} and PrimeComposer \cite{wangPrimeComposerFasterProgressively2024} do not provide their code-base or a publicly available implementation to evaluate their methods independently. The standard SD \cite{rombachHighResolutionImageSynthesis2022} inpainting method shipped with the model has already been evaluated in the TF-Icon \cite{luTFICONDiffusionBasedTrainingFree2023} paper and is outperformed by it. The same is true for Paint-by-Example \cite{yangPaintExampleExemplarbased2022}, which is already outperformed by AnyDoor \cite{chenAnyDoorZeroshotObjectlevel2023}. For generating a new background, we compare our method with ReplaceAnything \cite{chenReplaceAnythingYouWant2024} and the Background Replacement method inspired by a HuggingFace space provided by Shopify \cite{shopifyShopifyImageBackground2023}. 

\subsection{Benchmarks}
We evaluate our approach using two benchmark datasets. We derive the first one from the benchmark used by TF-Icon \cite{luTFICONDiffusionBasedTrainingFree2023}.
Following TF-Icon, we only use the Real-Real subset of their dataset to calculate quantitative metrics. We filter their initial dataset of 267 examples by removing samples that already contain an object to be replaced in the target image as this is not the task the SBR-method and our method are intended for. The filtered benchmark contains 209 samples.
The TF-Icon benchmark barely contains images from technical, design, and advertisement domains is, therefore, specifically suited to the TF-Icon use-case. Hence, we construct a second dataset to evaluate the capabilities of inserting technical and design products into new backgrounds. We use bicycle images from the BIKED dataset \cite{regenwetterBIKEDDatasetMachine2021}, car images from Stanford-Cars \cite{krause3DObjectRepresentations2013} and catalog-images of consumer products from Amazon-Berkeley-Objects \cite{collinsABODatasetBenchmarks2022} and Products10K \cite{baiProducts10KLargescaleProduct2020}. From each of the three categories, we select 20 samples randomly while manually labelling product types and color and ensuring product images are from different classes.
For insertion into existing backgrounds we generate backgrounds using SDXL and assign backgrounds to objects at random.
For the second task of inserting objects into a newly generated background, we only use the objects from our benchmark dataset and assign background prompts randomly.

\subsection{Metrics}
In our evaluation, we aim to assess the overall image quality and appeal of the resulting image composition, the alignment with the text-prompt and the geometric consistency of the inserted object compared to the ground-truth. We use the HPSv2-score \cite{wuHumanPreferenceScore2023} for overall image appeal, as it aims to replicate human preferences for natural and realistic images. To assess the alignment of the image composition with the text-prompt, we use the CLIP-score that measures the cosine similarity of the CLIP-embedded text and image \cite{radfordLearningTransferableVisual2021}. For geometric consistency, we use LPIPS \cite{zhangUnreasonableEffectivenessDeep2018}.
In addition to the automated metrics, we organize a user study with 15 participants to rate the overall image quality and appeal, the consistency with the text-prompt and the geometric consistency with the ground-truth of the composed image. We therefore randomly select 7 samples from each of our 3 benchmark categories and carry out the object-insertion task with the corresponding methods. We again compare our method to TF-Icon \cite{luTFICONDiffusionBasedTrainingFree2023} and AnyDoor \cite{chenAnyDoorZeroshotObjectlevel2023} for composition with an existing background with ReplaceAnything \cite{chenReplaceAnythingYouWant2024} and SBR \cite{shopifyShopifyImageBackground2023} for composition with a newly generated background.
To ensure objectivity, the human evaluation study is conducted blind with the shown examples being in random order.

\subsection{Composition into Existing Background}
 
\begin{figure*}[h]
\centering
\includegraphics[width=\linewidth]{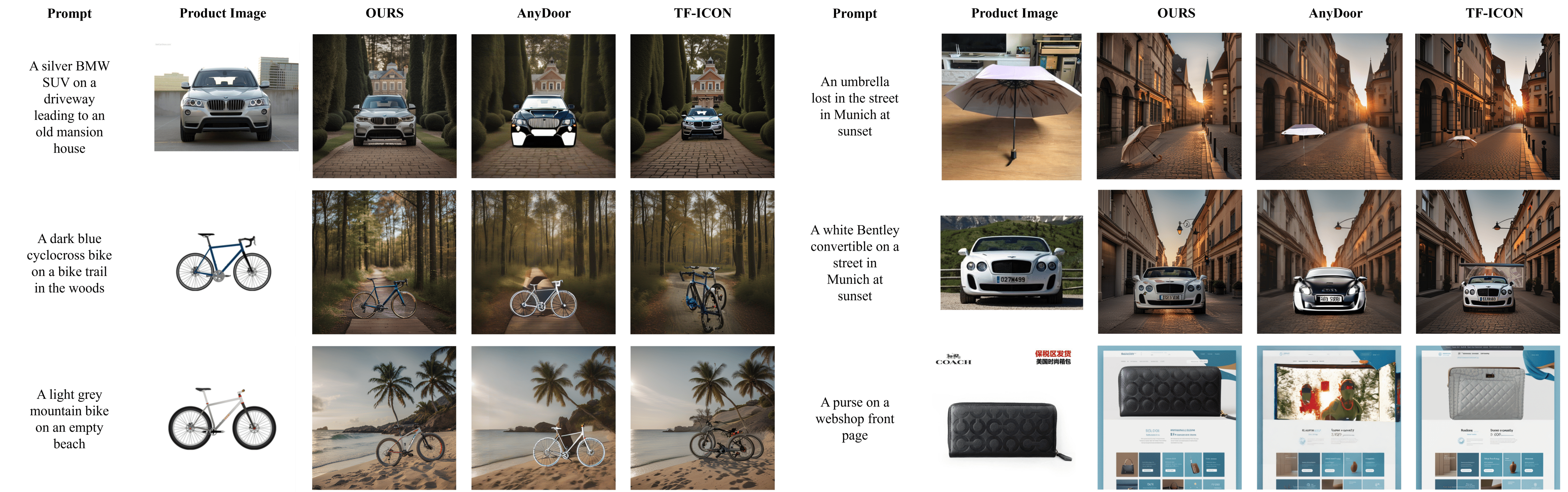}
\caption{{\bf Qualitative comparison with existing methods for insertion of product images into existing backgrounds}, including TF-Icon \cite{luTFICONDiffusionBasedTrainingFree2023} and AnyDoor \cite{chenAnyDoorZeroshotObjectlevel2023}. Our method improves seamless integration of the object into the background while preserving the geometry and structural integrity of the object.}
\label{fig:Compare_ExistingBG}
\vspace*{-0.25cm}
\end{figure*}

We compare our approach for inserting product images into existing backgrounds with the existing alternatives TF-Icon~\cite{luTFICONDiffusionBasedTrainingFree2023} and AnyDoor~\cite{chenAnyDoorZeroshotObjectlevel2023}. In ~\cref{fig:Compare_ExistingBG} we present some examples. The quantitative results are summarized in ~\cref{Metrics_ExistBG} and the results from our human evaluation study in ~\cref{HumanEval_ExistBG}.

In terms of image quality, our approach yields more appealing image compositions. The target object is embedded more realistically into the semantic framework of the existing background. The overall image appears more consistent. Our approach especially excels in visualizing fine-grained or beam-like structures. This is apparent with the bicycle images. Due to the adaptive masking approach, our method can handle empty spaces within an object and simultaneously preserve the structured geometry. While the quantitative evaluation using HPSv2 \cite{wuHumanPreferenceScore2023} moderately hints towards the supremacy of our approach, the human evaluation study strongly favors our approach over the alternatives.

For the alignment of the composed image with the text-description, InsertDiffusion outperforms the alternatives across all benchmarks and metrics. While TF-Icon \cite{luTFICONDiffusionBasedTrainingFree2023} is within range according to the CLIP-score, it falls noticeably short when evaluated by human annotators. AnyDoor \cite{chenAnyDoorZeroshotObjectlevel2023} does not allow for the formulation of prompt and therefore consistently achieves the lowest score for both CLIP and human evaluation. The most significant gap again exists for structural objects like bicycles. We suspect that this is due to the bicycle geometries being severely altered, sometimes even rendered unrecognizable, by the alternative methods. It is, therefore, difficult for human annotators to accurately evaluate the prompt alignment.

In analyzing the results for geometric consistency, we have to differentiate between the quantitative metrics and the human evaluation. For the quantitative LPIPS-score \cite{zhangUnreasonableEffectivenessDeep2018}, all three models achieve similar results. On its own benchmark, TF-Icon \cite{luTFICONDiffusionBasedTrainingFree2023} holds a 6\% advantage over InsertDiffusion. For the remaining datasets, the scores are almost identical. AnyDoor \cite{chenAnyDoorZeroshotObjectlevel2023} yields to better geometric consistency for the bicycle, car and consumer-product examples, according to the LPIPS-score. The human evaluation study paints a vastly different picture for geometric consistency. Averaged over our three benchmark sets, human annotators rate the geometric consistency of the inserted object better by a factor of 2.4 when using InsertDiffusion instead of AnyDoor \cite{chenAnyDoorZeroshotObjectlevel2023}. TF-Icon \cite{luTFICONDiffusionBasedTrainingFree2023} performs reasonably well overall, but is outperformed by our approach by a factor of 2.97 for the bicycle samples. We assume the reasons for the sharp difference between the quantitative results in LPIPS-score and the human evaluation to be twofold. The LPIPS-score measures the perceptual similarity of two images patches. It does not directly evaluate the structural composition of these patches and therefore most likely does not capture finer geometric details. The second reason has to do with the overall image quality and seamless integration of the object into its environment. We assume that the annotators in our human evaluation study generally prefer more realistic image compositions. With AnyDoor \cite{chenAnyDoorZeroshotObjectlevel2023}, the object often appears to have just been pasted onto the background, while many objects get significantly altered by TF-Icon \cite{luTFICONDiffusionBasedTrainingFree2023}.
\begin{table}[ht]
\caption{{\bf Comparison results} for the task of inserting an object into an existing background. "HPSv2", "CLIP" and "LPIPS" measure image appeal, text-alignment and geometric consistency. We evaluate the models on four different datasets, as described in Section \cref{experiments}.}
\scriptsize
\centering
\begin{tabularx}{\columnwidth}{ZZZZZ}
\toprule
{Dataset} & {Model} & {CLIP (↑)} & {HPSv2 (↑)} & {LPIPS (↓)} \\
\midrule
\multirow{3}{*}{TFI Benchmark}
    & \textit{TF-Icon} & 31.043 & 0.245 & \textbf{0.589} \\
    & \textit{AnyDoor} & 29.889 & 0.194 & 0.605  \\
    & \textit{\textbf{Ours}} & \textbf{31.801} & \textbf{0.250} & 0.624   \\
\multirow{3}{*}{Overall} 
    & \textit{TF-Icon} & 33.148 & 0.265 & 0.696 \\
    & \textit{AnyDoor} & 29.2982 & 0.224 & \textbf{0.652}   \\
    & \textit{\textbf{Ours}} & \textbf{34.997} & \textbf{0.287} & 0.699   \\
\multirow{3}{*}{Bikes}
    & \textit{TF-Icon} & 34.281 & 0.269 & 0.743  \\
    & \textit{AnyDoor} & 30.804 & 0.231 & \textbf{0.709} \\
    & \textit{\textbf{Ours}} & \textbf{36.058} & \textbf{0.286} & 0.757   \\
\multirow{3}{*}{Cars}
    & \textit{TF-Icon} & 33.121 & 0.287 & 0.679  \\
    & \textit{AnyDoor} & 27.637 & 0.230 & \textbf{0.654}   \\
    & \textit{\textbf{Ours}} & \textbf{34.979} & \textbf{0.310} & 0.672   \\
\multirow{3}{*}{Products}
    & \textit{TF-Icon} & 31.987 & 0.238 & 0.647  \\
    & \textit{AnyDoor} & 29.453 & 0.213 & \textbf{0.621}   \\
    & \textit{\textbf{Ours}} & \textbf{33.955} & \textbf{0.267} & 0.642 \\
\bottomrule
\end{tabularx}
\label{Metrics_ExistBG}
\vspace*{-0.25cm}
\end{table}

\begin{table}[ht]
\caption{{\bf Human Evaluation} for the task of inserting an object into an existing background. We measure overall image appeal, alignment of the composed image with the text-prompt and the geometric consistency of the inserted object on a scale of 1 (non-existent) to 5 (perfect).}
\scriptsize
\centering
\begin{tabularx}{\columnwidth}{ZZZZZ}
\toprule
{Dataset} & {Model} & {Overall \newline Appeal (↑)} & {Prompt \newline Alignment (↑)} & {Geometric \newline Consistency (↑)} \\
\midrule
\multirow{3}{*}{Overall} 
    & \textit{TF-Icon} & 2.780 & 3.752 & 3.333 \\
    & \textit{AnyDoor} & 1.905 & 2.762 & 1.695 \\
    & \textit{\textbf{Ours}} & \textbf{3.410} & \textbf{3.790} & \textbf{3.790}   \\
\multirow{3}{*}{Bikes}
    & \textit{TF-Icon} & 1.711 & 2.033 & 1.156 \\
    & \textit{AnyDoor} & 2.038 & 2.714 & 2.114 \\
    & \textit{\textbf{Ours}} & \textbf{3.211} & \textbf{3.900} & \textbf{3.433}   \\
\multirow{3}{*}{Cars}
    & \textit{TF-Icon} & 2.781 & 3.752 & 3.333 \\
    & \textit{AnyDoor} & 1.905 & 2.762 & 1.695 \\
    & \textit{\textbf{Ours}} & \textbf{3.410} & \textbf{3.790} & \textbf{3.790}  \\
\multirow{3}{*}{Products}
    & \textit{TF-Icon} & 2.743 & 3.248 & 1.895 \\
    & \textit{AnyDoor} & 2.343 & 2.895 & 2.257 \\
    & \textit{\textbf{Ours}} & \textbf{3.571} & \textbf{4.162} & \textbf{3.695} \\
\bottomrule
\end{tabularx}
\label{HumanEval_ExistBG}
\vspace*{-0.25cm}
\end{table}


\subsection{Composition with Generated Background}

To compare our approach with alternatives for inserting the product-image representations into generated backgrounds, we use the same evaluations as in the previous section. The quantitative results and the results from the human evaluation are summarized in \cref{Metrics_NewBG} and \cref{HumanEval_NewBG}. Since AnyDoor \cite{chenAnyDoorZeroshotObjectlevel2023} and TF-Icon \cite{luTFICONDiffusionBasedTrainingFree2023} do not provide for the option of novel background generation, we compare our approach to ReplaceAnything \cite{chenReplaceAnythingYouWant2024} and SBR \cite{shopifyShopifyImageBackground2023}. Both methods are specifically designed to generate new backgrounds for the object to be inserted into.

\begin{figure*}[h]
\centering
\includegraphics[width=\linewidth]{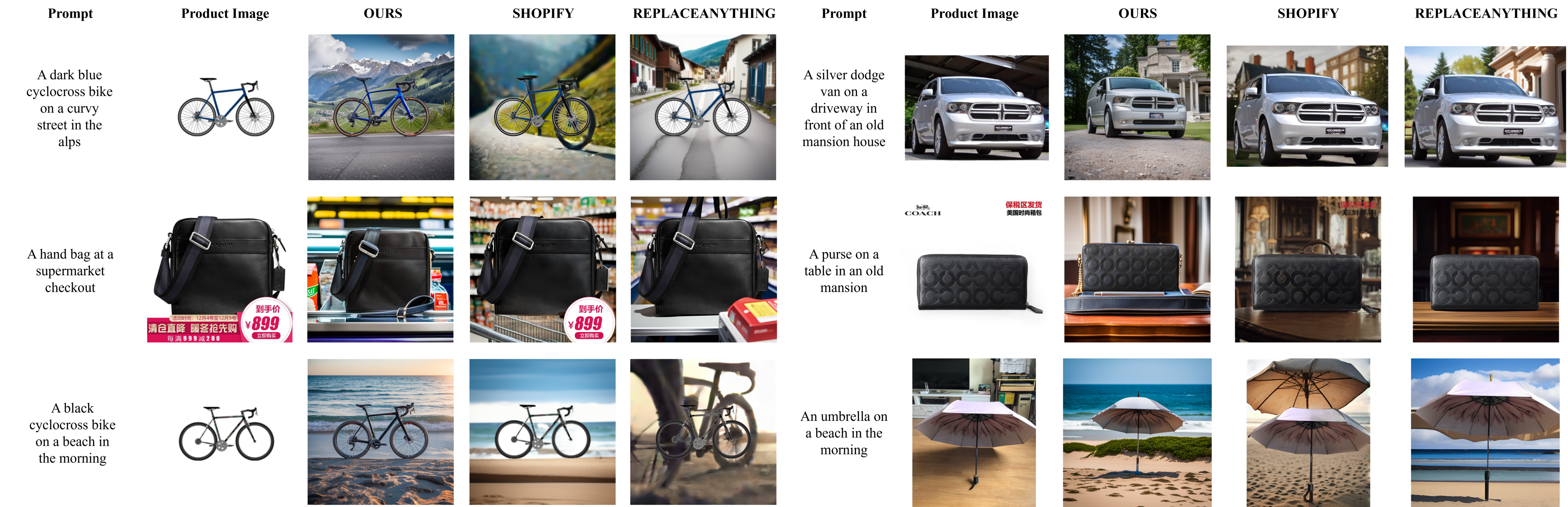}
\caption{{\bf Qualitative comparison with existing methods for insertion of product images into newly generated backgrounds}, including ReplaceAnything and SBR. Our method composes the object and the background in a more natural manner, being able to adapt the object to fit the background seamlessly while preserving its key geometric and semantic characteristics.}
\label{fig:Compare_NewBG}
\vspace*{-0.25cm}
\end{figure*}

In the quantitative analysis, our approach achieves superior results for human preference (HPSv2) and alignment with the text description (CLIP). On our benchmark dataset composed of bicycle, car and product images, we surpass ReplaceAnything by 8,67\%  and SBR by 16,60 \% for the human preference score. For the CLIP-score, our approach holds an 8,50\% advantage over ReplaceAnything and 11,39\% over SBR. In terms of geometric consistency, ReplaceAnything performs best according to the LPIPS-score. The object geometries in the composed image are more consistent with the original. However, this consistency comes at the cost of sacrificing the quality of the overall image composition. As shown in \cref{fig:Compare_NewBG}, some objects appear to be simply cut out and pasted onto the new background. This is especially true for bicycle images. In masking the objects adaptively and then allowing for marginal modifications, the compositions of our method look more realistic and seamless.
The human evaluation study supports this conclusion. While the geometric consistency of ReplaceAnything is preferred by the evaluators, our approach appears to generate vastly more appealing image compositions. Over the entire benchmark dataset, we achieve a 28,13\% better evaluation score.

\begin{table}[h]
\caption{{\bf Comparison results} for the task of inserting an object into a newly generated background. "HPSv2", "CLIP" and "LPIPS" measure image appeal, text-alignment and geometric consistency. We evaluate the models on the three categories of our benchmark dataset, as described in Section \cref{experiments}.}
\scriptsize
\centering
\begin{tabularx}{\columnwidth}{ZZZZZ}
\toprule
{Dataset} & {Model} & {CLIP (↑)} & {HPSv2 (↑)} & {LPIPS (↓)} \\
\midrule
\multirow{3}{*}{Overall} 
    & \textit{ReplaceAnything} & 31.070 & 0.265 & \textbf{0.244} \\
    & \textit{SBR} & 30.264 & 0.247 & 0.435  \\
    & \textit{\textbf{Ours}} & \textbf{33.710} & \textbf{0.288} & 0.403   \\
\multirow{3}{*}{Bikes}
    & \textit{ReplaceAnything} & 33.275 & 0.277 & \textbf{0.213}  \\
    & \textit{SBR} & 31.449 & 0.245 & 0.527 \\
    & \textit{\textbf{Ours}} & \textbf{33.810} & \textbf{0.296} & 0.474   \\
\multirow{3}{*}{Cars}
    & \textit{ReplaceAnything} & 28.722 & 0.285 & \textbf{0.280}  \\
    & \textit{SBR} & 28.914 & 0.270 & 0.438   \\
    & \textit{\textbf{Ours}} & \textbf{33.837} & \textbf{0.301} & 0.305   \\
\multirow{3}{*}{Products}
    & \textit{ReplaceAnything} & 30.977 & 0.236 & \textbf{0.242}  \\
    & \textit{SBR} & 30.430 & 0.227 & 0.341   \\
    & \textit{\textbf{Ours}} & \textbf{33.483} & \textbf{0.268} & 0.430 \\
\bottomrule
\end{tabularx}
\label{Metrics_NewBG}
\end{table}

\begin{table}[h]
\caption{{\bf Human Evaluation} for the task of inserting an object into a new background. We measure overall image appeal, alignment of the composed image with the text-prompt and the geometric consistency of the inserted object on a scale of 1 (non-existent) to 5 (perfect).}
\scriptsize
\centering
\begin{tabularx}{\columnwidth}{ZZZZZ}
\toprule
{Dataset} & {Model} & {Overall Appeal (↑)} & {Prompt Alignment (↑)} & {Geometric Consistency (↑)} \\
\midrule
\multirow{3}{*}{Overall} 
    & \textit{ReplaceAnything} & 3.057 & 3.600 & \textbf{4.140} \\
    & \textit{SBR} & 2.016 & 2.686 & 3.397 \\
    & \textit{\textbf{Ours}} & \textbf{3.917} & \textbf{3.905} & 3.498  \\
\multirow{3}{*}{Bikes}
    & \textit{ReplaceAnything} & 1.657 & 2.914 & \textbf{3.486} \\
    & \textit{SBR} & 1.610 & 2.467 & 2.762 \\
    & \textit{\textbf{Ours}} & \textbf{4.181} & \textbf{3.971} & 3.419  \\
\multirow{3}{*}{Cars}
    & \textit{ReplaceAnything} & 3.762 & 3.971 & \textbf{4.610} \\
    & \textit{SBR} & 2.152 & 2.829 & 3.619 \\
    & \textit{\textbf{Ours}} & \textbf{3.990} & \textbf{4.086} & 3.971  \\
\multirow{3}{*}{Products}
    & \textit{ReplaceAnything} & \textbf{3.743} & \textbf{3.914} & \textbf{4.324} \\
    & \textit{SBR} & 2.286 & 2.762 & 3.810 \\
    & \textit{\textbf{Ours}} & 3.705 & 3.752 & 3.210  \\
\bottomrule
\end{tabularx}
\label{HumanEval_NewBG}
\end{table}


\subsection{Ablations}
To verify the efficiency of our approach and evaluate its limitations, we perform a number of ablations. For the components of the InsertDiffusion architecture, we compare different model versions available within Diffusers. We also investigate the influence of the last refinement step compared to just performing the masking and inpainting operations until the intermediate image composition. We further evaluate if increased human interference in the generative pipeline leads to more appealing image compositions. Finally, we discuss ablations of the image colorization option.\\

\begin{figure}[h]
\centering
\includegraphics[width=\linewidth]{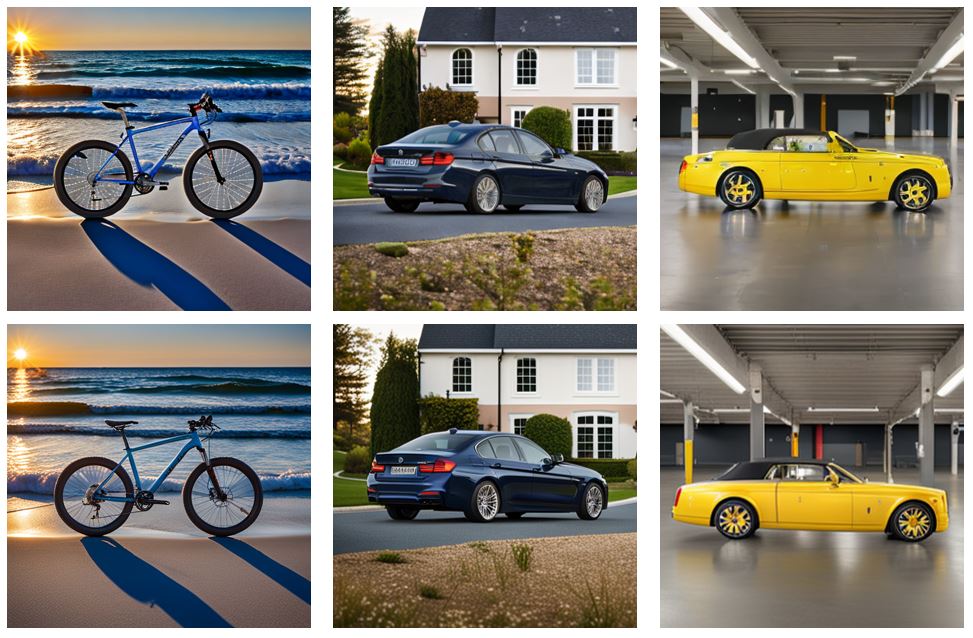}
\caption{{\bf InsertDiffusion with and without SDXL refinement.} Top: Intermediate image compositions (without refinement). Bottom: Image compositions with refinement.}
\label{fig:Refinement}
\end{figure}

{\bf InsertDiffusion Architecture.} 
A significant ablation to the architecture is to leave out the refinement step where the intermediate image composition is noised and then denoised again using SDXL. To verify the usefulness of this additional step, we evaluate the results of our architecture with and without the refinement step on our benchmark dataset as well as on the TF-Icon benchmark. The results are summarized in \cref{tab:ablation_no_refinement}. Although relatively small, the quantitative analysis confirms the impression from the visual inspection (\cref{fig:Refinement}) of the image compositions. The final images are more appealing when using the additional refinement step and show increased consistency with the text description. The intermediate compositions show an increased geometric consistency for some cases. This is somewhat expected since the refinement step adds noise to the object and then denoises it solely based on the guidance from the text-prompt.
\begin{table}[h]
\caption{{\bf Ablation for refinement stage}. We compare our method against our method without the refinement stage. Refining the image after inpainting improves image quality and prompt alignment.}
\scriptsize
\centering
\begin{tabularx}{\columnwidth}{>{\centering\arraybackslash}p{2cm}>{\centering\arraybackslash}p{1.5cm}>{\centering\arraybackslash}p{1.5cm}>{\centering\arraybackslash}p{1.5cm}}
\toprule
{Dataset} & {Metric} & {Ours} & {\begin{tabular}[c]{@{}c@{}}Ours \\ w\textbackslash o refinement\end{tabular}}\\
\midrule
\multirow{3}{*}{TF-Icon Benchmark} 
    & \textit{HPSv2 (↑)} & 0.250 & 0.230 \\
    & \textit{CLIP (↑)} & 31.801 & 30.729 \\
    & \textit{LPIPS (↓)} & 0.623 & 0.611\\
\multirow{3}{*}{Overall}
    & \textit{HPSv2 (↑)} & 0.287 & 0.275 \\
    & \textit{CLIP (↑)} & 34.997 & 34.030 \\
    & \textit{LPIPS (↓)} & 0.699 & 0.701 \\
\bottomrule
\end{tabularx}
\label{tab:ablation_no_refinement}
\end{table}

\begin{figure}[h]
\centering
\includegraphics[width=\linewidth]{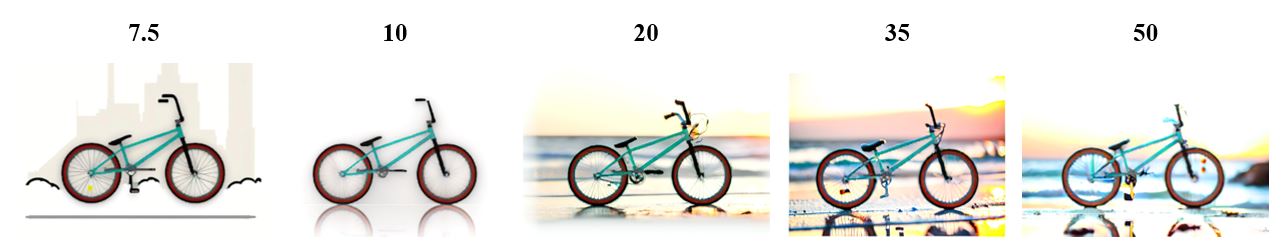}
\caption{{\bf Using SDXL for inpainting} at various guidance scales. Note
that the image always contains a lot of white. Increasing the guidance leads to reduced object quality and consistency.}
\label{fig:SDXL_Inpaint}
\end{figure}

Our modularized architecture (see \cref{fig:Architecture}) allows for the utilization of different versions of models from Diffusers. Usually, one would prefer the most recent model versions as they posses the most advanced generative capabilities. For our architecture however, we find that using SDXL with the available inpainting function leads to worse results than using SD-2.1. For the intermediate composition, the generated backgrounds only contain faint and abstract structures. This is most likely an issue with the inpainting implementation. Furthermore, we observe that increasing the guidance scale reduced the quality and conistency of the inserted object, as shown in \cref{fig:SDXL_Inpaint}. Using SDXL to synthesize the intermediate image composition does not allow for realistic visualizations of the object.\\
\vspace{-0.25cm}

{\bf Additional User Interference.} 
We investigate the effect of giving the user additional control by allowing to chose between 5 variations of the intermediate and the final image compositions. When object colorization is required, the user can also chose between 5 samples. We again evaluate the results quantitatively on our benchmark dataset. This ablation was also included in the humamn evaluation study. The results are summarized in \cref{tab:Ours_vs_Interact}. While the additional interference has no significant impact on the overall appeal and the alignment with the text description, it does seem to have a small impact in improving the geometric consistency, as both LPIPS and the human evaluation score slightly prefer the interactive method over the standard approach.

\begin{table}[h]
\caption{{\bf Ablation for User Interference.} We compare our standard method against our method with increased user interference.}
\scriptsize
\centering
\begin{tabularx}{\columnwidth}{>{\centering\arraybackslash}p{3cm}>{\centering\arraybackslash}p{2cm}>{\centering\arraybackslash}p{2cm}}
\toprule
{Metric} & {Ours} & {Ours (interactive)}  \\
\midrule
\textit{HPSv2 (↑)} & 0.288 & 0.292 \\
\textit{CLIP (↑)} & 33.710 & 33.431 \\
\textit{LPIPS (↓)} & 0.403 & 0.388  \\
\textit{Overall Appeal (↑)} & 3.917 & 3.819 \\
\textit{Prompt Alignment (↑)} & 3.905 & 4.086 \\
\textit{Geometric Consistency (↑)} & 3.498 & 3.717 \\
\bottomrule
\end{tabularx}
\label{tab:Ours_vs_Interact}
\end{table}

\begin{figure}[h]
\centering
\includegraphics[width=\columnwidth]{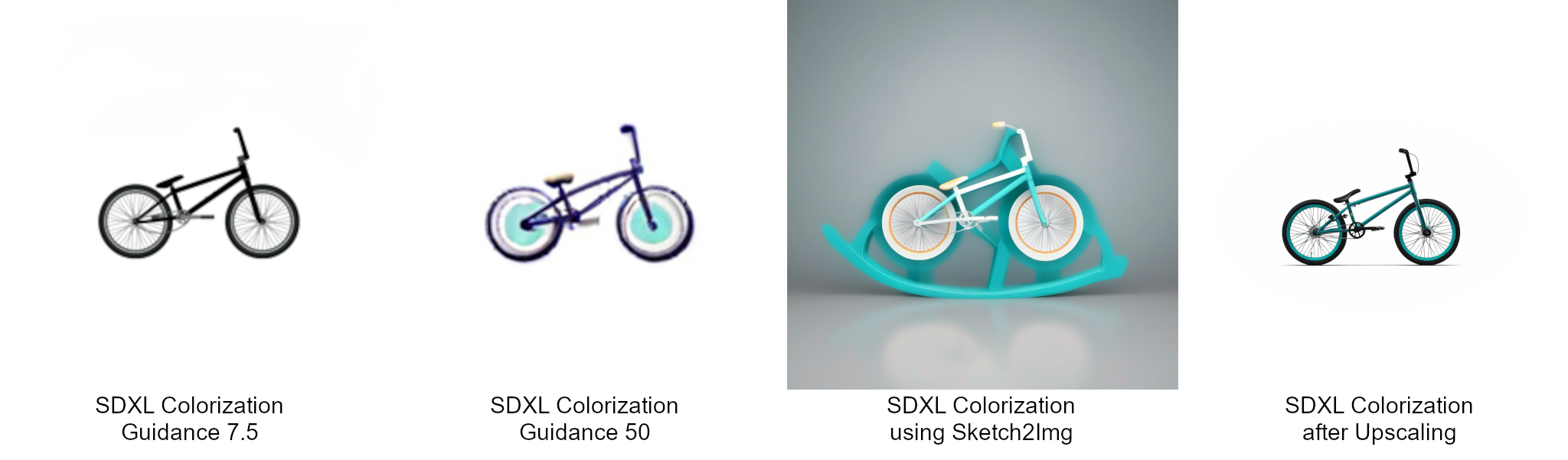}
\caption{{\bf Ablation for colorization without upscaling.} Standard masked img2img diffusion with SDXL does not colorize the image, increasing the guidance until color becomes visible destroys the structure, sketch2img diffusion does more than colorization.}
\label{fig:colorization_upscaling}
\vspace*{-0.25cm}
\end{figure}

{\bf Colorization.} For the optional colorization, we compare using a masked SDXL, SDXL together with a ControlNet Sketch-to-Image adapter \cite{zhangAddingConditionalControl2023} and SDXL colorization after upsampling the input image to increase the resolution and clarity of the object geometry (see \cref{fig:colorization_upscaling}). We observe that without upsampling, low guidance leads to the object not being colorized at all while high guidance leads to unwanted parts within the image being colorized. The ControlNet adapter that treats the b/w image as an input sketch does not colorize the geometry properly. We find that upsampling the input image before passing it to a masked SDXL image-to-image transformation colorizes the geometry reliably.

\section{Limitations and Future Work}
A primary limitation of our approach is its dependence on adequate scaling and positioning of the inserted object by the user. Our model cannot automatically detect where to place the object in a given background. The limited modifications we carry out on the background image and the object do not prevent unrealistic scenarios if the object is misplaced. Prime examples are objects that appear to be floating in the air.
In addition, since we utilize pretrained latent diffusion models and do not modify or train them, our approach is fundamentally limited by their generative capabilities. For example, with the current selection of models, we can not accurately generate or maintain text within images. Further limitations are the semantic and geometric consistency of the inserted objects. Due to the final refinement step, the inserted objects are inevitably altered.

To overcome the limitations, future research may explore approaches to ensure the consistency of the inserted objects by extracting targeted image features of the original object and injecting them into the final refinement step.
Another direction for future experiments is using more powerful diffusion models, such as SDXL, for inpainting in the architecture.
Adding additional preprocessing steps for the object and using specifically trained or finetuned LDMs are directions worth exploring for domain-specific applications.

\section{Societal Impact}
Our approach can enable improved visualizations in technical design processes and potentially lead to more user-centered experiences in digital product marketing. By eliminating the need for additional training and finetuning, we provide an efficient and sustainable method with a low skill barrier to create creative product visualizations.

However, this does not come without potential risks. It can be misused to create fake images of real objects, thereby contributing to misinformation or deception, which is the case for many state-of-the-art models for image generation \cite{rostamzadehEthicsandCreativity2021}. The model is designed for the visualization of technical products but might still be used to create unethical content or violate privacy by placing someone's personalized objects in compromising situations. With the current selection of diffusion models, our method does not allow to accurately place a human in new backgrounds, as the human face get altered by the final refinement step. We acknowledge that this can become another risk with future diffusion models or if models are implemented that have specifcally been finetuned to reproduce human faces.  Additionally, the automation of tasks in content creation might affect jobs in fields like photography, grapic design or marketing. By being based on SD, InsertDiffusion  may inadvertently amplify biases present in the training data of SD \cite{rombachHighResolutionImageSynthesis2022, EsserNoteOnBiases2020}.

\section{Conclusion}
In this work, we present InsertDiffusion, a training-free architecture based on state-of-the-art diffusion models for inserting objects into given or novel backgrounds. By adaptively masking the target area for the inserted object and using a step-wise combination of inpainting and image-to-image transformations, we achieve seamless compositions with the backgrounds and are able to visualize the objects in realistic scenes. Our approach outperforms alternative methods in terms of image quality and alignment with textual descriptions and achieves state-of-the-art object consistency. It excels at visualizing technical and CAD-like images. The modular architecture allows for the potential integration of more capable diffusion models as soon as they become available and is aimed to be easy-to-use by utilizing models from the diffusers library on HuggingFace.

{\small
\bibliographystyle{ieee_fullname}
\bibliography{InsertDiffusion}
}

\end{document}